# Rethinking the Role of Collaborative Robots in Rehabilitation*

Vivek Gupte
Idiap Research Institute & EPFL
Switzerland
vivek.gupte@idiap.ch

Shalutha Rajapakshe
Idiap Research Institute & EPFL
Switzerland
shalutha.rajapakshe@idiap.ch

Emmanuel Senft
Idiap Research Institute
Switzerland
esenft@idiap.ch

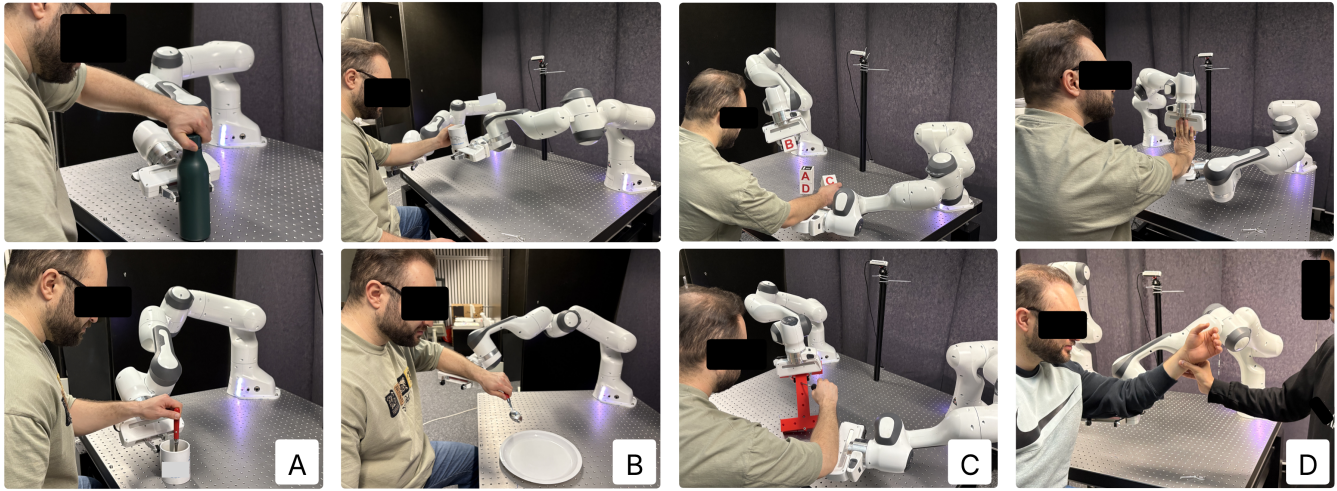

Figure 1: Collaborative robots (cobots) can play a wide range of roles in a rehabilitation workspace. Cobots can A) conduct task-oriented exercises by co-manipulating objects, B) support movement when practicing daily tasks, C) develop interactive exercises to evoke a sense of accomplishment, and D) help manage therapist effort and time by assisting with tasks such as weight bearing.

## Abstract

Current research on collaborative robots (cobots) in physical rehabilitation largely focuses on repeated motion training for people undergoing physical therapy (PuPT), even though these sessions include phases that could benefit from robotic collaboration and assistance. Meanwhile, access to physical therapy remains limited for people with disabilities and chronic illnesses. Cobots could support both PuPT and therapists, and improve access to therapy, yet their broader potential remains underexplored. We propose extending the scope of cobots by imagining their role in assisting therapists and PuPT before, during, and after a therapy session. We discuss how cobot assistance may lift access barriers by promoting ability-based therapy design and helping therapists manage their time and effort. Finally, we highlight challenges to realizing these roles, including advancing user-state understanding, ensuring safety, and integrating cobots into therapists' workflow. This view opens new research questions and opportunities to draw from the HRI community's advances in assistive robotics.

*This work was funded by Swiss National Science Foundation under Grant No. 10002419

## CCS Concepts

• **Human-centered computing** → **Collaborative interaction**; • **Computer systems organization** → *Robotic autonomy*.

## Keywords

Rehabilitation, Assistive robots, Collaborative robots, Accessibility

## 1 Introduction

People with disabilities and chronic conditions often rely on physical therapy to maintain movement and function. Effective therapy involves frequent, high-intensity, and repetitive exercises [22]. Yet, many people face insufficient access to therapy or shortened and infrequent sessions partly due to therapists facing heavy caseloads [28]. As a result, people undergoing physical therapy (PuPT) may struggle to build and sustain their functional gains.

Rehabilitation robotics offers promising tools to assist physiotherapists. Wearable exoskeletons and robotic manipulators can deliver intense exercises with precise repetitions while supporting an individual's joints [1, 14, 26]. Such robots can reduce therapist workload while increasing exercise volume and improving motor control, muscle strength, and joint coordination [2, 22, 29]. But custom exoskeletons might be too expensive or complex.

Collaborative robots (cobots) have recently been explored as rehabilitation robots [22]. Cobots have joint level force and torque sensing that enables safe and dynamic physical interactions, and their mass producibility makes them more affordable than custom exoskeletons. In rehabilitation settings, cobots can assist in limb movements [22], accommodate individual exercise needs [3], and provide user-adaptive assistance [21]. Their use has primarily focused on limb-movement exercises.

However, a typical therapy session involves far more than repeating movement-based exercises. For example, it includes warm-up and stretching to prepare for exercises, and task-oriented training to practice reaching, grasping, and handling everyday objects [16]. Therapists may facilitate movement completion, stabilize parts of the limb, correct compensatory behavior, or walk with PuPT while supporting their weights [4, 10, 12]. They often spend considerable time and effort on setting up the equipment and transferring people with mobility impairments between machines [9, 24].

Cobots, with their sensing, manipulation, and physical interaction capabilities may support therapists and PuPT across several rehabilitation tasks. By taking on physically demanding and time-consuming tasks, they might reduce therapist's efforts and give them greater control over their time. Thus, they have the potential to improve therapy accessibility and improve therapists' job quality, a crucial but often overlooked objective [23].

To better understand the role of cobots in rehabilitation, we held an introductory discussion with a group of physiotherapists who reflected on how cobots might fit into their practice. Informed by their insights and a review of current cobots usage in rehabilitation, this report proposes an expanded set of directions for cobots beyond traditional movement-based exercises. Specifically, we explore how cobots could:

- assist therapists and PuPT before, during, and after physical therapy sessions;
- improve access to therapy by promoting personalization and ability-based therapy design; and
- support a broader set of physical interaction modalities.

## 2 Current Role of Cobots in Rehabilitation

Upper-limb rehabilitation robots are typically classified as exoskeletons and end-effector robots. While exoskeletons are wearable robots that connect to a person's limbs at multiple points, end-effector robots physically connect to a single, often distal, point on the limb [22]. Exoskeletons provide support to the PuPT's arm joints and improve range of motion, but are often custom built, expensive, and need an adjusted fit [14]. End-effector robots encourage three-dimensional movements, multi-joint coordination and fine-motor control without customized fitting [14, 30], but provide limited joint-level support which can be critical in early rehabilitation [14].

Cobots, developed for safe and proximate human-robot interaction, have been used as end-effector type rehabilitation robots. They are tethered to PuPT's distal limb through handles or wearable attachments and follow fixed trajectories [17, 20, 25]. These trajectories are encoded using tools like virtual fixtures [20, 30] and are often learned from therapist demonstrations [8]. Cobots can also provide different levels of assistance using the 'assist-as-needed' (AAN) paradigm, varying assistance level based on performance, tracking accuracy, or biomechanical and physiological measurements [7, 21, 25]. Their use has expanded with the rise of medically certified robotic platforms such as the KUKA LBR Med robot [11].

Beyond single-arm setups, recent work in upper-limb rehabilitation has explored bimanual cobots to simultaneously exercise both of the user's arms [30]. These robots have been used in leader–follower configurations, where a therapist-side leader modulates assistance transmitted to the user through a follower [7]. They have also been used asymmetrically, with one robot stabilizing the user's proximal upper-limb and the other assisting movements [5].

Despite these efforts, there have been limited attempts in exploring a cobot's role beyond movement-based exercises. For instance, few studies have used cobots to provide therapist-like assistance [18], as objects in the environment (such as doors and drawers) with which the user interacts [6], or for co-manipulating objects with the user [8]. Even then, the primary role of a cobot is to support direct movement of the user's limbs.

While efforts to encode complex movements and develop assist-as-needed algorithms are important, much of the focus remains on repetitive limb movements for predefined exercises. However, we believe that cobots can offer far more opportunities for assisting therapists and PuPT alike.

## 3 Extending the Role of Cobots in Rehabilitation

To understand the various roles cobots could play in a rehabilitation setting, we invited five physiotherapists to our lab for an interactive group discussion. We introduced them to cobots and demonstrated cobot functionalities through simple exercises programmed on the Franka Emika Panda robot arm. While interacting with the robot, they highlighted common challenges they face in their practice and we discussed how cobots could be helpful as assistants.

This research was approved by Idiap's Data and Research Ethics Committee under project REHABOT. Initially intended as an introductory meeting between the therapists and the researchers, our discussion uncovered interesting themes surrounding cobot roles in rehabilitation. Therefore, we obtained post-hoc written consent from the physiotherapists to report these insights.

Drawing from our discussion and a review of existing literature on cobots in rehabilitation, we outline four ways cobots could support rehabilitation (1) during therapy sessions, by assisting movement and task practice; (2) before and after sessions, by supporting set-up, warm-up, tear-down, and assessments; (3) by improving access to therapy; and (4) through versatile physical interaction modalities.

### 3.1 During a Physical Therapy Session

Physical therapy involves PuPT performing different types of exercises often supported by a therapist who conducts the session. The exercises include moving the arms at different joints, reaching for, handling, and manipulating objects of different sizes and shapes, and practicing activities of daily living (ADLs).

*3.1.1 Robots for Supporting Movement.* Repetitive limb motion is important in therapy, and robots have shown success in assisting with such movements. However, in addition to guiding PuPT's limbs, therapists also support their movement by protecting against tissue impingement [10], managing load across proximal joints [4], and preventing substitutional movement patterns that could inhibit recovery [12]. While most robots can perform repetitive movements by tethering to the PuPT, they can rarely support them in ways similar to the therapists.

Cobots' ability to be in different configurations while maintaining the same end-effector pose together with joint-level compliance, may allow cobots to interact with different parts of the PuPT's limb and torso without restricting their movements. Cobots may stabilize the PuPT's proximal limb, protect against tissue impingement, or stop compensatory movements using end-effectors or other links. While such interactions have been explored for point-to-point movement exercises (see [5]), they have not been extended to more complex ADL tasks to enable effective progression towards multi-joint coordination and skill transfer. Exploring these approaches might allow the therapists to use cobots in exercises with increasing complexity, as PuPT progress through therapy.

Cobots could also be employed during activities like stretching and warm-up which are common but tailored to PuPT's needs. They are different from other repetitive movement exercises in that they might have time-varying amplitudes, complex resistance profiles, and varying frequencies depending on the user's muscle spasticity. Cobots may assist in active or passive stretching, or stabilize the therapy seeker's limbs and torso as they warm-up and changing between stiff and compliant configurations may provide PuPT with flexible fixtures to 'stretch against'.

Additionally, robots might provide support, for example, by reducing the weight felt by the PuPT and allowing more independent complex movements. Exploring movement-based exercises where the PuPT's hands are not always tethered to the robot may reveal new ways to provide assistance during therapy. Assisting the PuPT in this way might free their hands to practice meaningful tasks, such as using spoons or turning bottle caps (see Fig. 1).

*3.1.2 Robots as Interactive Exercise Environment.* Task-oriented training involves reaching, grasping or handling everyday objects of varying size and shape, like doors, cups, or taps. These activities mimic ADLs that help PuPT apply movements learned during therapy in everyday lives. In these tasks, the robot might play the role of the object itself. This idea has been explored to an extent (see [6]), by using the robot to mimic the forces exerted by a door. The robot may adapt the difficulty of handling these 'robotic' objects by changing its controller parameters at the therapist's discretion, or by estimating the individual's ability.

During task-oriented training, therapists often place different objects on a table as per the intended exercise difficulty and help the PuPT in handling these objects by moving them closer or further away, or by replacing them after a repetition. However, after initial teaching and practice, the session often turns repetitive as the therapists spend time resetting the environment, helping PuPT handle objects, or support movements as they complete these tasks. Collaborative robots in such scenarios may play a dual role: acting as assistants that reset the workspace and as rehabilitation robots that support the PuPT's limbs. As assistants, robots may reset the workspace, adapt task difficulty by changing the object placement, and support movements as detailed in Section 3.1.1.

Furthermore, co-manipulation tasks such as performing assembly might be administered as exercises. In these tasks, the robot might help or challenge the PuPT in accomplishing certain aspects of the tasks, while carrying out aspects of the task that the PuPT is unable to perform. These co-manipulation tasks could even be gamified providing users with entertainment and accomplishment as well as motivation to continue therapy [27]. Using robots in this way may give time back to the therapists, which they might choose to spend monitoring the PuPT, or otherwise.

*3.1.3 Robots as Assistants to the Therapists.* Therapists often need to support the PuPT's limb weight during therapy such as when demonstrating a new finger or wrist exercise. In such situations, therapists may invest more effort in supporting the limbs diverting their attention from the exercise itself, or recruit another therapist for support. Collaborative robots might be used to support most of PuPT's weight in this scenario while allowing therapists to freely interact with the PuPT without over-exerting themselves and with more control. Thus, collaborative robots might be envisioned as assistants to the therapists, acting as an extra arm to help stabilize and hold the PuPT.

## 3.2 Before and After a Physical Therapy Session

*3.2.1 Set-up and Teardown of Equipment.* Therapists often spend significant amounts of time in setting up and tearing down the exercise equipment before and after the session respectively. Saving their time and effort during this period might help improve quality of their work. Cobots acting as 'all-in-one' exercise machines might allow quick and automatic changes in the set-up from one exercise to next. Further, these robots may allow personalizing the exercise to each user with minimal effort from the therapists.

*3.2.2 Assessments, Benchmarking, and Evaluation.* Cobots might be promising as an assessment and evaluation tool for the therapists. Cobots can empower the therapists to quickly change parameters and tailor exercises to the PuPT's needs, or quickly benchmark the exercises and gather insights about varying abilities of the PuPT. Cobots can leverage sensed forces during an interaction to provide useful feedback about the interaction to the therapists, track the PuPT's progress, and help plan the therapy session. Thus, combining the ability to quickly change exercise parameters with the ability to evaluate PuPT performance could help therapists make well-informed decisions. In addition, sharing progress over time with the PuPT might be a source of encouragement for them [28].

## 3.3 Cobots for Improving Access to Therapy

*3.3.1 Disability-centered Approach.* People with disabilities and chronic illnesses rely on physical therapy for managing their symptoms, enabling independence, and promoting participation in daily activities. Yet, these individuals often face numerous barriers towards accessing therapy making it difficult to benefit from it [28]. Using cobots in physical therapy might enable personalization and ability-based therapy design, helping lift prevalent access barriers.

**Ability-based Therapy Design using Cobots.** Physical therapy outcomes for people with disabilities must align with their unique needs and personal goals [15], which differ from person to person. Cobots hold potential to tackle these variations by functioning as versatile exercise machines, and by playing assistive and rehabilitative roles during a session. For example, an individual who faces difficulty maintaining balance might have to perform exercises while lying down. While conventional exercise machines can rarely accommodate such needs, cobots with their more flexible workspace can help. They might enable ability-based therapy design by allowing therapists to easily tune robot control parameters and select from a wide range of robot behaviors.

**Lifting Physiological Barriers to Therapy.** People with disabilities have fluctuating abilities and 'good' or 'bad' days that might discourage or prevent them from participating in therapy. For instance, this might be due to different levels of fatigue or pain that they feel right after exercising leading to a known physiological access barrier [28].

Cobots may infer an individual's present ability by sensing forces during the physical interaction and adapt their behavior according to the user's needs. They can dynamically vary assistance based on the individuals current ability and comfort level to help them complete range-of-motion exercises or practice ADLs. Such information about the PuPT's state might also be useful for therapists to update session goals or prescribe suitable activities. Thus, cobots in physical therapy might improve access to therapy.

*3.3.2 Improving Accessibility by Assisting Therapists.* Insufficient physiotherapy appointments are another barrier to accessing therapy, often attributed to limited availability of physiotherapists. This barrier might be partly lifted by using collaborative robots to help therapists reclaim their time and manage effort. While not a part of a therapy session, therapists spend considerable time and effort on assistive activities such as supporting movement between exercise machines for PuPT with mobility impairments, or helping PuPT with dressing. We can draw from research on assistive robots for people with disabilities to develop robots for support with mobility, transfer, and dressing [19], helping manage therapist effort and time. By studying the role of collaborative robots towards this goal, we might not only take steps towards making therapy more accessible, but possibly improve the therapist's job quality. Collaborative robots playing a multifaceted role of assistive and rehabilitation robots might be key to realizing this goal.

## 3.4 Physical Interaction Modalities

As we imagine the scope of cobots in robot-assisted rehabilitation, it might be worth considering the different ways they might interface with the PuPT. Most existing works use handles that require the PuPT to grasp onto the robot's end-effector [17], which might be ineffective in exercises where the PuPT's hand and fingers need to be free. In addition, PuPT such as stroke survivors grasping onto a handle might promote spasticity and slow down progress. As such, collaborative robots with easy-to-switch and exercise-specific end-effectors which could get into configurations that allow easier grips while performing exercises.

To keep the user's hand free while supporting their limb weight, cobot end-effector could be attached to PuPT's wrists [20] or to the back of their hands [13, 25] using orthotic attachments. These attachments might need to maintain extension in the PuPT's palm and fingers to avoid spasticity. Hence, approaches such as elastic gloves with pneumatic actuation as cobot attachments could provide a universal fit and a fine-grained extension control to gradually help PuPT out of muscle spasticity.

## 4 Conclusion and Future Work

Presently, most research on cobots in rehabilitation focuses on movement-based interactions where PuPT's hands are tethered to the cobot end-effector. However, cobots in rehabilitation could play a multifaceted role in assistive and collaborative tasks. In this report we highlighted aspects of therapy where cobots could be useful before, during and after the session. However, bringing these new cobot roles to reality requires more work in the following:

**Advancing user state understanding:** Extending the use of cobots to support and stabilize the limb during task-oriented movement based exercises will require a nuanced understanding of the person's physiological state to determine assistive actions, and an accurate estimation of their spatial state to identify where support is required. Providing dynamic assistance in untethered co-manipulation tasks might require advances in user state perception using modalities such as vision and understanding user preferences, comfort, and needs through interactive feedback. These capabilities must be developed using non-invasive, privacy preserving approaches to tracking and storing user state.

**Ensuring safe and robust proximate interactions:** As robots and humans might work in close proximity in these applications, the robots must maintain safe and robust behavior. This might require efforts in understanding contextual safety requirements, the user's abilities, and their biomechanics. It also calls for robust controllers that can handle varying physical interactions and allow behavior customization.

**Achieving smooth workflow integration:** Cobots deployed in rehabilitation centers must integrate smoothly within the existing workflows and not disrupt them. Setting up the robots for exercises, modifying robot behavior, assistance levels, compliance, and correcting robot movements will require intuitive interfaces for therapists and PuPT. Developing such behaviors and interfaces will require close collaboration with therapists and PuPT to capture and communicate the requirements of physical interaction effectively. Successful robot integration will require positive user-perceptions, improvements to therapist's job quality, and the cost effectiveness of the robot platform.

Imagining broader roles for cobots in rehabilitation opens new research questions and opportunities to draw from advances made by the assistive robotics and physical human-robot interaction communities. Through our planned future interactions with therapists and PuPT, we plan on investigating more deeply, the extent to which cobots could address issues faced in physical rehabilitation.